\documentclass[nonacm, sigconf]{acmart}

\renewcommand\footnotetextcopyrightpermission[1]{} 
\usepackage{float}
\usepackage{booktabs} 
\usepackage{times}
\usepackage{graphicx}
\usepackage{amsmath}
\usepackage{enumitem}
\usepackage{hyperref}
\DeclareMathSizes{12}{30}{16}{12}
\setcopyright{rightsretained}
\acmDOI{ }
\acmISBN{}
\acmYear{2018}
\copyrightyear{2018}
\acmArticle{}
\acmPrice{}

\newcommand\blfootnote[1]{%
  \begingroup
  \renewcommand\thefootnote{}\footnote{#1}%
  \addtocounter{footnote}{-1}%
  \endgroup
}

\begin{document}
\title{Context-Aware Attention for Understanding Twitter Abuse}

\author{Tuhin Chakrabarty}
\affiliation{%
  \institution{Columbia University}
  \city{New York}
  \state{NY}
  \postcode{10027}
}
\email{tc2896@columbia.edu}

\author{Kilol Gupta}
\affiliation{%
  \institution{Columbia University}
  \city{New York}
  \state{NY}
  \postcode{10027}
}
\email{kilol.gupta@columbia.edu}


\begin{abstract}
\blfootnote{The full published version of this work is available at: \url{https://www.aclweb.org/anthology/W19-3508/}. Please use the published version for citation purposes.}The original goal of any social media platform is to facilitate users to indulge in healthy and meaningful conversations. But more often than not, it has been found that it becomes an avenue for wanton attacks. We want to alleviate this issue and hence we try to provide a detailed analysis of how abusive behavior can be monitored in Twitter. The complexity of the natural language constructs makes this task challenging. We show how applying contextual attention to Long Short Term Memory networks help us give near state of art results on multiple benchmarks abuse detection data sets from Twitter.
\end{abstract}


\maketitle

\section{Introduction \& Related Work}
Any social interaction involves an exchange of viewpoints and thoughts. But these views and thoughts can be caustic. Often we see that users resort to verbal abuse to win an argument or overshadow someone's opinion. On Twitter, people from every sphere have experienced online abuse. Be it a famous celebrity with millions of followers or someone representing a marginalized community such as LGBTQ, Women and more. We want to channelize Natural Language Processing (NLP) for social good and aid in the process of flagging abusive tweets and users. Detecting abuse on Twitter can be challenging, particularly because the text is often noisy. Abuse can also have different facets. \cite{waseem2016hateful} released one of the initial data sets from Twitter with the goal of identifying what constitutes racism and sexism.
\cite{davidson} in their work pointed out that hate speech is different from offensive language and released a data set of 25k tweets with the goal of distinguishing hate speech from offensive language.
\begin{table}[H]
\begin{center}
\begin{tabular}{|l|l|}
\hline
\multicolumn{2}{|l|}{\begin{tabular}[c]{@{}l@{}}Stop saying dumb blondes with pretty\\ faces as you need a pretty face to pull \\ them off !!! \#mkr\end{tabular}} \\ \hline
\multicolumn{2}{|l|}{\begin{tabular}[c]{@{}l@{}}In Islam women must be locked in their \\ houses and Muslims claim this is treating\\ them well\end{tabular}} \\ \hline
\end{tabular}
\caption{Tweets from \cite{waseem2016hateful} data set demonstrating online abuse}
\end{center}
\end{table}

They find that racist and homophobic tweets are more likely to be classified as hate speech but sexist tweets are generally classified as offensive. \cite{golbeck} introduced a large, hand-coded corpus of online harassment data for studying the nature of harassing comments and the culture of trolling. Keeping these motivations in mind, we make the following salient contributions:
\begin{itemize}
	\item We build a deep context-aware attention-based model for abusive behavior detection on Twitter . To the best of our knowledge ours is the first work that exploits context aware attention for this task.
	\item Our model is robust and achieves consistent performance gains in all the three abusive data sets
	\item We show how context aware attention helps in focusing  on certain abusive keywords when used in specific context and improve the performance of  abusive behavior detection  .
\end{itemize}

\section{Related Work}
Existing approaches to abusive text detection can be broadly divided into two categories: 1) Feature intensive machine learning algorithms such as Logistic Regression (LR), Multilayer Perceptron (MLP) and etc. 2) Deep Learning models which learn feature representations on their own. \cite{waseem2016hateful} released the popular data set of 16k tweets annotated as belonging to sexism, racism or none class \footnote{http://github.com/zeerakw/hatespeech  \label{footnote 1}}, and provided a  feature engineered model  for detection of abuse in their corpus. \cite{davidson} use a similar handcrafted feature engineered model to identify offensive language and distinguish it from  hate speech. \cite{badjatiya2017deep} in their work, experiment with multiple deep learning architectures for the task of hate speech detection on Twitter using the same data set by \cite{waseem2016hateful}. Their best-reported F1-score is achieved using Long Short Term Memory Networks (LSTM) + Gradient Boosting.

On the data set released by \cite{waseem2016hateful}, \cite{park2017one} experiment with a two-step approach of detecting abusive language first and then classifying them into specific types i.e. racist, sexist or none. They achieve best results using a Hybrid Convolution Neural Network (CNN) with the intuition that character level input would counter the purposely or mistakenly misspelled words and made-up vocabularies. \cite{pavlopoulos2017deeper} in their work ran experiments on the Gazetta dataset and the DETOX system (\cite{wulczyn2017ex}) and show that a Recurrent Neural Network (RNN) coupled with deep, classification-specific attention outperforms the previous state of the art in abusive comment moderation. In their more recent work \cite{pavlopoulos2017improved} explored how user embeddings, user-type embeddings, and user type biases can improve their previous RNN based model on the Gazetta dataset. Attentive neural networks have been shown to perform well on a variety of NLP tasks (\cite{yang2016hierarchical}, \cite{yinattention}). \cite{yang2016hierarchical} use hierarchical contextual attention for text classification (i.e attention both at word and sentence level) on six large scale text classification tasks and  demonstrate that the proposed architecture outperform previous methods by a substantial margin.We primarily focus on word level attention because most of the tweets are single sentence tweets.

\section{Model}
The best choice for modeling tweets was Long Short Term Memory Networks (LSTMs) because of their ability to capture long-term dependencies by introducing a gating mechanism that ensures the proper gradient propagation through the network. We use bidirectional LSTMs because of their inherent capability of capturing information from both: the past and the future states.
A bidirectional LSTM (BiLSTM) consists of a forward LSTM $\overrightarrow{f}$ that reads the sentence from $x_{1}$ to $x_{T}$ and a backward LSTM $\overleftarrow{f}$ that reads the sentence from $x_{T}$ to $x_{1}$, where T is the number of words in the sentence under consideration and $x_{i}$ is the $i_{th}$ word in the sentence. We obtain the final annotation for a given word $x_{i}$, by concatenating the annotations from both directions (Eq. [1]). \cite{speech} show that LSTMs can benefit from depth in space.Stacking multiple recurrent hidden layers on top of each other, just as feed forward layers are stacked in the conventional deep networks give performance gains .And hence we choose stacked LSTM for our experiments.

\begin{center}
\includegraphics[scale=0.50]{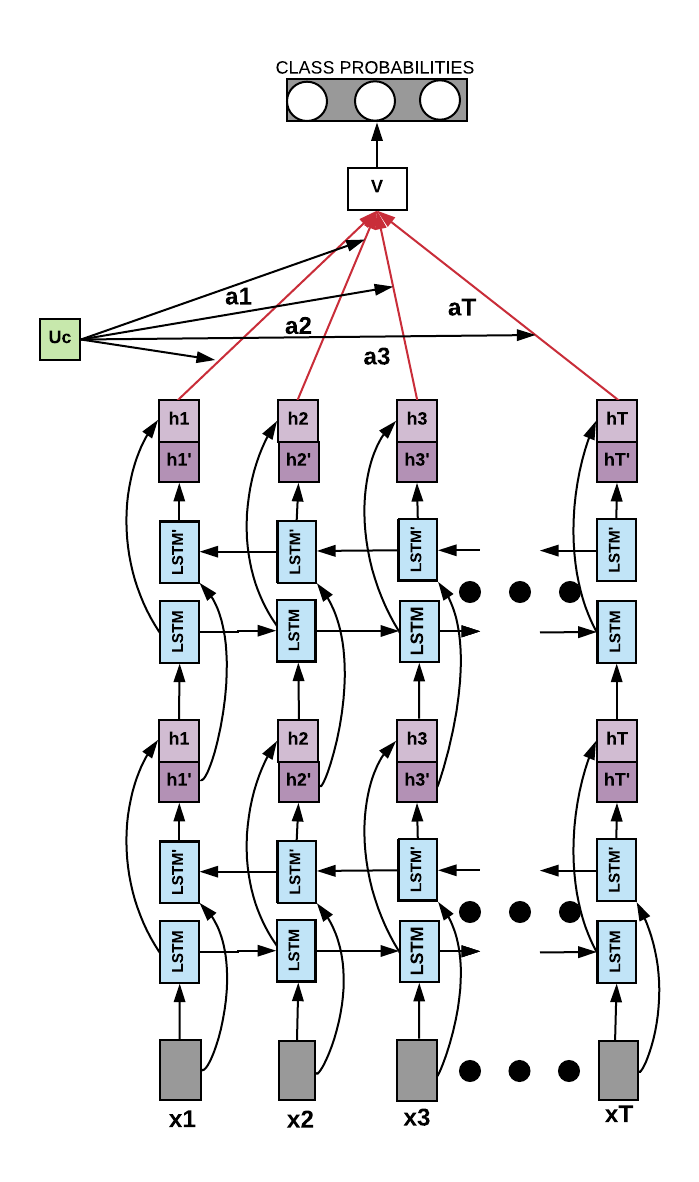}
\captionof{figure}{Architecture for Stacked BiLSTM + Word Level Contextual Attention. Figure is inspired by \cite{yang2016hierarchical}}
\end{center}

\subsection{Word Attention}
The attention mechanism assigns a weight to each word annotation that is obtained from the BiLSTM layer. We compute the fixed representation \textbf{\Large v} of the whole message as a weighted sum of all the word annotations which is then fed to a final fully-connected Softmax layer to obtain the class probabilities.
We first feed the LSTM output $h_{i}$ of each word $x_{i}$ through a  Multi Layer Perceptron to get $u_{i}$ as its hidden representation. $u_{c}$ is our word level context vector that is randomly initialized and learned as we train our network. Once $u_{i}$ is obtained we calculate the importance of the word as the similarity of $u_{i}$ with $u_{c}$  and get a normalized importance weight $\alpha_{i}$ through a softmax function. The context vector $u_{c}$ can be seen as a tool which filters which word is more important over all the words like that used in the LSTM. Figure 2 shows the high-level architecture of this model. $W_{h}$ and $b_{h}$ are the attention layer’s weights and biases. More formally,

\begin{equation}
\begin{aligned}
    h_{i} = [\overrightarrow{f}, \overleftarrow{f}]
    \forall{i = 1,...T}
\end{aligned}
\end{equation}

\begin{equation}
\begin{aligned}
    u_{i} =tanh(W_{h}.h_{i} + b_{h})
\end{aligned}
\end{equation}
\begin{equation}
    a_{i} = \frac{ u_{i}^{T}u_{c} }{ \sum_{j=1}^{T} u_{j}^{T}u_{c} }
\end{equation}
\begin{equation}
    v = \sum_{i=1}^{T}a_{i}h_{i}
\end{equation}

\section{Experiments}
In this section  we talk about data sets first and then go on to show our results obtained on these three data sets .We also show some examples where our model failed . Finally we show how attention helps us understand the model in a better fashion.

\subsection{Data Sets}
We have used the 3 benchmark data sets for abusive content detection on Twitter.
At the time of the experiment, the \cite{waseem2016hateful} data set had a total of \textbf{15,844} tweets out of which \textbf{1,924} were labelled as belonging to racism, \textbf{3,058} as sexism and \textbf{10,862} as none. The \cite{davidson} data set had a total of \textbf{25,112} tweets out of which \textbf{1498} were labelled as hate speech, \textbf{19,326} as offensive language and \textbf{4,288} as neither. For the \cite{golbeck} data set, there were \textbf{20,362} tweets out of which \textbf{5,235} were positive harassment examples and \textbf{15,127} were negative. 

We call \cite{waseem2016hateful} data set as D1 , \cite{davidson} data set as D2 and \cite{golbeck} as D3
\begin{table}
\begin{center}
 \begin{tabular}{||c |c ||} 
 \hline
 \textbf{Data Set} & \textbf{Tweets Count} \\ [0.5ex] 
 \hline\hline
 \cite{waseem2016hateful} & 15,844 \\ 
 \hline
 \cite{davidson} & 25,112\\
 \hline
 \cite{golbeck} & 20,362 \\
 \hline
\end{tabular}
\caption{Data sets and their total tweets count}
\label{table:1}
\end{center}
\end{table}

For tweet tokenization, we use \textbf{Ekphrasis} which is a text processing tool built specially from social platforms such as Twitter. 

\cite{baziotis2017datastories} use a big collection of Twitter messages (330M) to generate word embeddings, with a vocabulary size of 660K words, using GloVe (\cite{pennington2014glove}). We use these pre-trained word embeddings for initializing the first layer (embedding layer) of our neural networks.

\subsection{Results}
The network is trained at a learning rate of 0.001 for 10 epochs, with a dropout of 0.2 to prevent over-fitting. The results are averaged over 10-fold cross-validations for D1 and D3 and 5 fold cross-validations for D2 because \cite{davidson} reported results using 5 fold CV. Because of class imbalance in all our data sets, we report weighted F1 scores.

Table 3 shows our results in detail. We compare our model with the best models reported in each paper. Because \cite{golbeck} is a data set paper, we cannot fill the corresponding row. \textbf{*} denotes the numbers from baseline papers. All the results were reproducible except for the one marked red. For (Waseem and Hovy, 2016) data set,  (Badjatiyaet al., 2017) claim that using Gradient Boosting with LSTM embeddings obtained from random word embeddings boosted their performance by 12 F1 from 81.0 to 93.0. When we tried to reproduce the result, we did not find any significant improvement over 81. Results show that our model is robust when it comes to the performance on all of the three data sets. 

\begin{table}[H]
\begin{center}
\begin{tabular}{|p{2.2cm}|p{0.8cm}|p{0.8cm}|p{0.8cm}|}
\hline
 {\textbf{Models}}& {\textbf{D1}} & {\textbf{D2}} & {\textbf{D3}} \\ \hline\hline
{\cite{waseem2016hateful}} & 73.8* & 82.3 & 63.0 \\ \hline
{\cite{davidson}} & 78.0 & 90.0* & 69.0  \\ \hline
{\cite{golbeck}}& - & - & -  \\ \hline
{\cite{park2017one}} & 82.7* & 88.0 & 70.6  \\ \hline
{\cite{badjatiya2017deep}} & \textbf{\textcolor{red}{93.1}*} & 88.0 & 65.7 \\ \hline\hline
{\textbf{Our Model}} & 84.2 & \textbf{91.1} & \textbf{72.7} \\ \hline
\end{tabular}
\caption{Data sets and the results of different models. We reproduced the results for each model on three of the data sets.}
\end{center}
\end{table}
We also share some examples from the three data sets in Figure \ref{fig:fig2} which our BiLSTM attention model could not classify correctly. On closer investigation we find that most cases where our model fails are instances where annotation is either noisy or the difference between classes are very blurred and subtle.

\begin{center}
\includegraphics[scale=0.33]{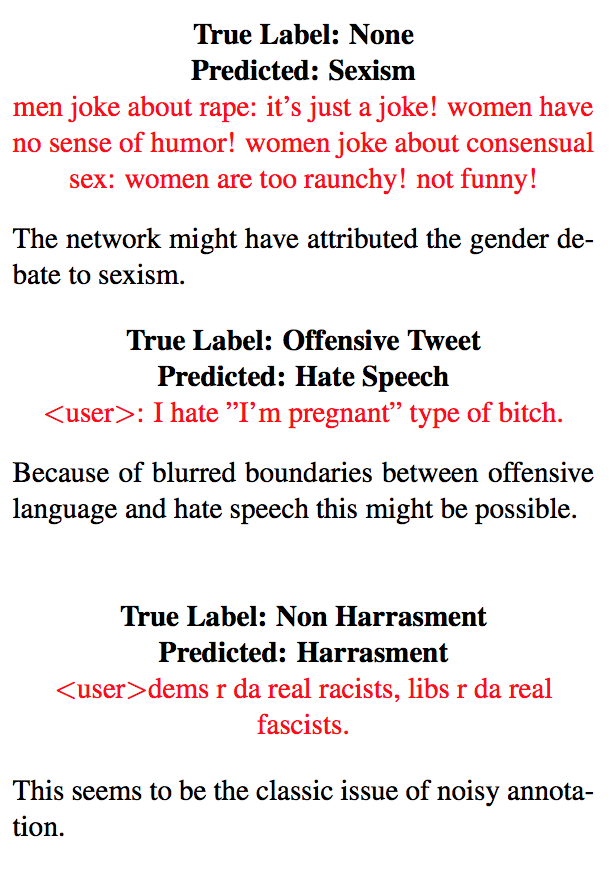}
\captionof{figure}{ The first tweet is a  tweet from \cite{waseem2016hateful}, the second tweet is a tweet from from \cite{davidson} data set and the third from the \cite{golbeck} datset }
\label{fig:fig2}
\end{center}

\subsection{Why Contextual Attention?}
Attention mechanism enables our neural network to focus on the relevant parts of the input more than the irrelevant parts while performing a 
prediction task. But the relevance is often dependant on the context and so the importance of words is highly context dependent. For example, the word \textbf{islam} may appear in the realm of Racism as well as in any normal conversation.The top tweet in Figure \ref{fig:fig3} belongs to \textbf{None} class while the bottom tweet belongs to \textbf{Racism} class.

\begin{center}
\includegraphics[scale=0.20]{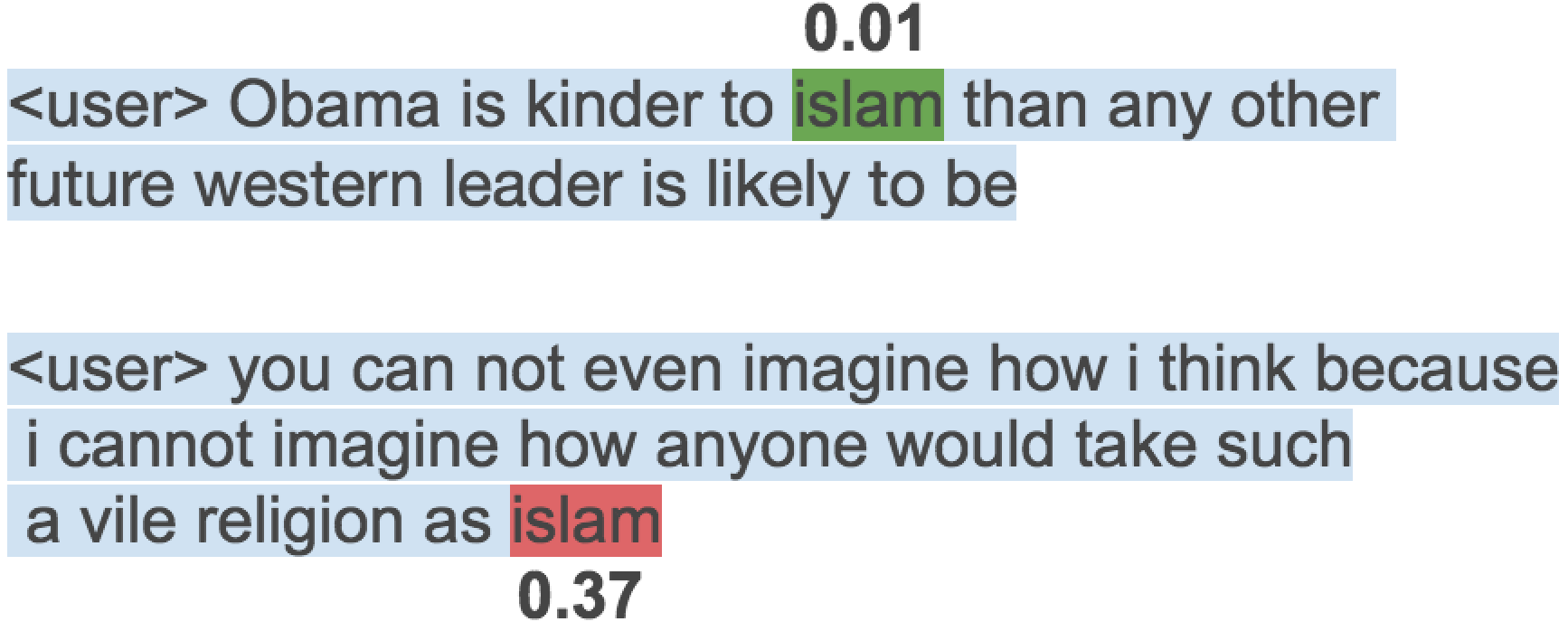}
\captionof{figure}{An example showing how our model captures
diverse context and assigns context-dependent weights to the same word in two different tweets.}
\label{fig:fig3}
\end{center}

\subsection{Attention Heat Map Visualization}
The color intensity corresponds to the weight given to each word by the contextual attention.
\begin{center}
\includegraphics[scale=0.15]{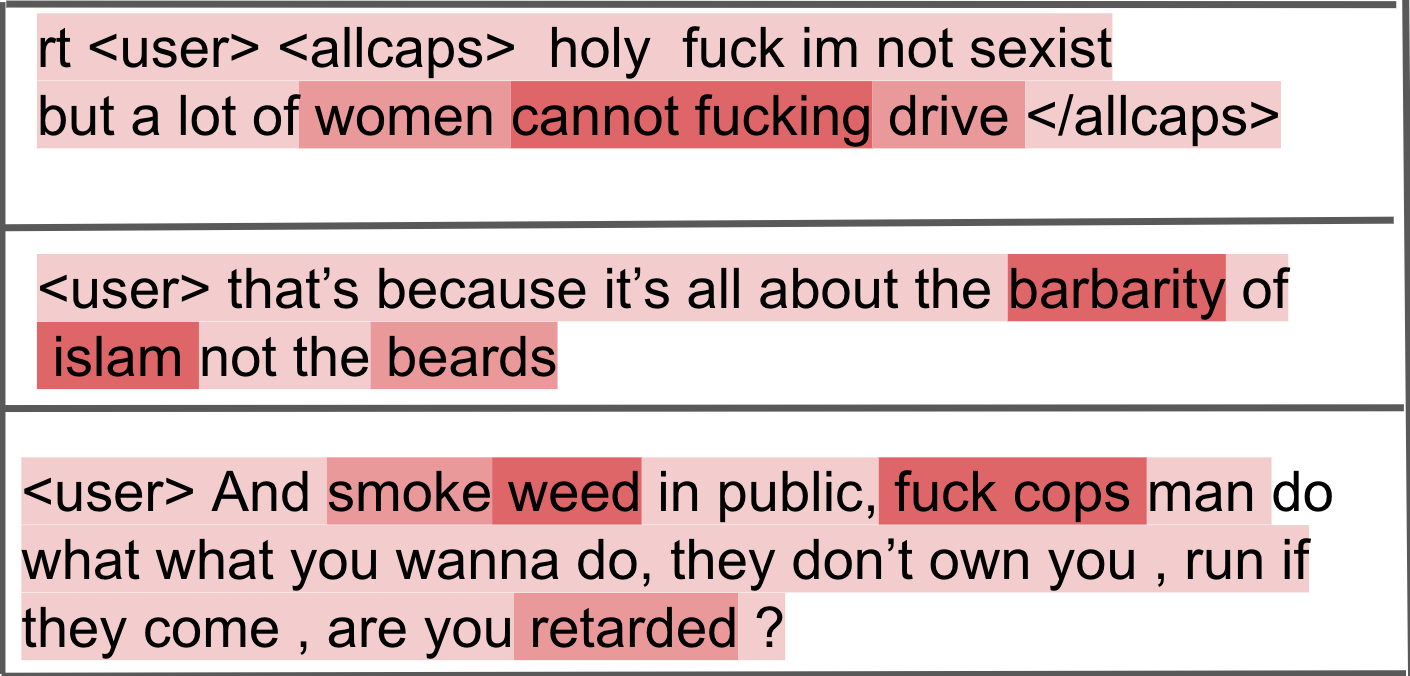}
\captionof{figure}{ The first tweet is a sexist tweet from \cite{waseem2016hateful} where as the second tweet is an example of racist tweet from the same datset . The third tweet is from \cite{davidson} data set labelled as offensive language.}
\label{fig:fig4}
\end{center}

\section{Conclusion and Future Work}
We successfully built a deep context-aware attention-based model and applied it to the task of abusive tweet detection. We ran experiments on three relevant data sets and empirically showed how our model is robust when it comes to detecting abuse on Twitter. We also show how context-aware attention helps us to interpret the model's performance by visualizing the attention weights and conducting thorough error analysis.\\
As for future work, we want to experiment with a model that learns user embeddings from their historical tweets. We also want to model abusive text classification in Twitter by taking tweets in context because often standalone tweets don't give a clear picture of a tweet's intent.

\bibliographystyle{ACM-Reference-Format}
\bibliography{sample-bibliography}

\end{document}